\documentclass[conference]{IEEEtran}
\IEEEoverridecommandlockouts
\usepackage{cite}
\usepackage{amsmath,amssymb,amsfonts}
\usepackage{algorithmic, comment}
\usepackage{graphicx}
\usepackage{textcomp}
\usepackage{balance}
\usepackage{xcolor}
\usepackage{adjustbox}
\usepackage{multirow}
\usepackage{wrapfig}
\usepackage{subfig}
\usepackage[normalem]{ulem}
\useunder{\uline}{\ul}{}
\usepackage{cuted}

\def\BibTeX{{\rm B\kern-.05em{\sc i\kern-.025em b}\kern-.08em
    T\kern-.1667em\lower.7ex\hbox{E}\kern-.125emX}}
\begin{document}

\title{Is Facial Recognition Biased at Near-Infrared Spectrum As Well?\\
}

\author{\IEEEauthorblockN{Anoop Krishnan\textsuperscript{\textsection}, Brian Neas\textsuperscript{\textsection} and Ajita Rattani}
\IEEEauthorblockA{\textit{School of Computing} \\
\textit{Wichita State University}\\
Wichita, USA\\
\{axupendrannair, bcneas\}@shockers.wichita.edu, ajita.rattani@wichita.edu}
}

\maketitle
\begingroup\renewcommand\thefootnote{\textsection}
\footnotetext{Equal contribution. Authors are mentioned in the alphabetical order of their last name.}
\endgroup

\begin{abstract}
   Published academic research and media articles suggest face recognition is biased across demographics. Specifically, unequal performance is obtained for women, dark-skinned people, and older adults. However, these published studies have examined the bias of facial recognition in the visible spectrum (VIS). Factors such as facial makeup, facial hair, skin color, and illumination variation have been attributed to the bias of this technology at VIS. The near-infrared (NIR) spectrum offers an advantage over VIS in terms of robustness to factors such as illumination changes, facial make-up, and skin color. 
       
        Therefore, it is worth-while to investigate the bias of the facial recognition at near-infrared spectrum~(NIR).
       This first study investigates the bias of face recognition system at NIR spectrum. To this aim, two popular NIR facial image datasets namely, CASIA-Face-Africa and NotreDame-NIVL consisting of African and Caucasian subjects, respectively, are used to investigate the bias of facial recognition technology across gender and race. Interestingly, experimental results suggest equitable performance of the face recognition across gender and race at NIR spectrum.
       
    
\end{abstract}


\section{Introduction}
\textit{"São Paulo subway ordered to suspend use of facial recognition"}~\cite{news}, this is one among the several media articles published at a well-known press that 
suggested facial recognition technology is biased across demographics.  Apart from media articles, several published academic studies on the bias of face recognition and visual attribute classification algorithms (such as gender-, age-classification and BMI prediction) also suggest performance differences across gender, race, and age~\cite{DBLP:journals/cviu/BeveridgeGPD09, 8756625, DBLP:journals/tifs/KlareBKBJ12, 9660182,GIVENS2013236, DBLP:journals/pami/Best-RowdenJ18,DBLP:conf/fat/BuolamwiniG18,DBLP:journals/corr/abs-2207-10246, DBLP:conf/icmla/KrishnanAR20, Siddiqui_2022_CVPR}. 
An unfair (biased) algorithm is one whose decisions are skewed towards a particular group of people. 

Specifically, these studies suggested that face recognition is less accurate for females and people of color~\cite{DBLP:journals/tifs/KlareBKBJ12,DBLP:journals/pami/Best-RowdenJ18,DBLP:conf/aies/RajiB19,DBLP:conf/cvpr/Vera-RodriguezB19,DBLP:conf/wacv/AlbieroSVZKB20,DBLP:conf/fgr/OTooleAPD11}. In December $2019$, a Face Recognition Vendor Test (FRVT) report by the National Institute of Standards and Technology (NIST)~\cite{42061} confirmed the differential accuracy of face recognition systems across race, gender, and age-based demographic groups on a dataset of $8.49$ million people using $189$ mostly commercial algorithms from $99$ developers.

\begin{figure}[tbh!]{}
\centering
\includegraphics[width=0.85\linewidth]{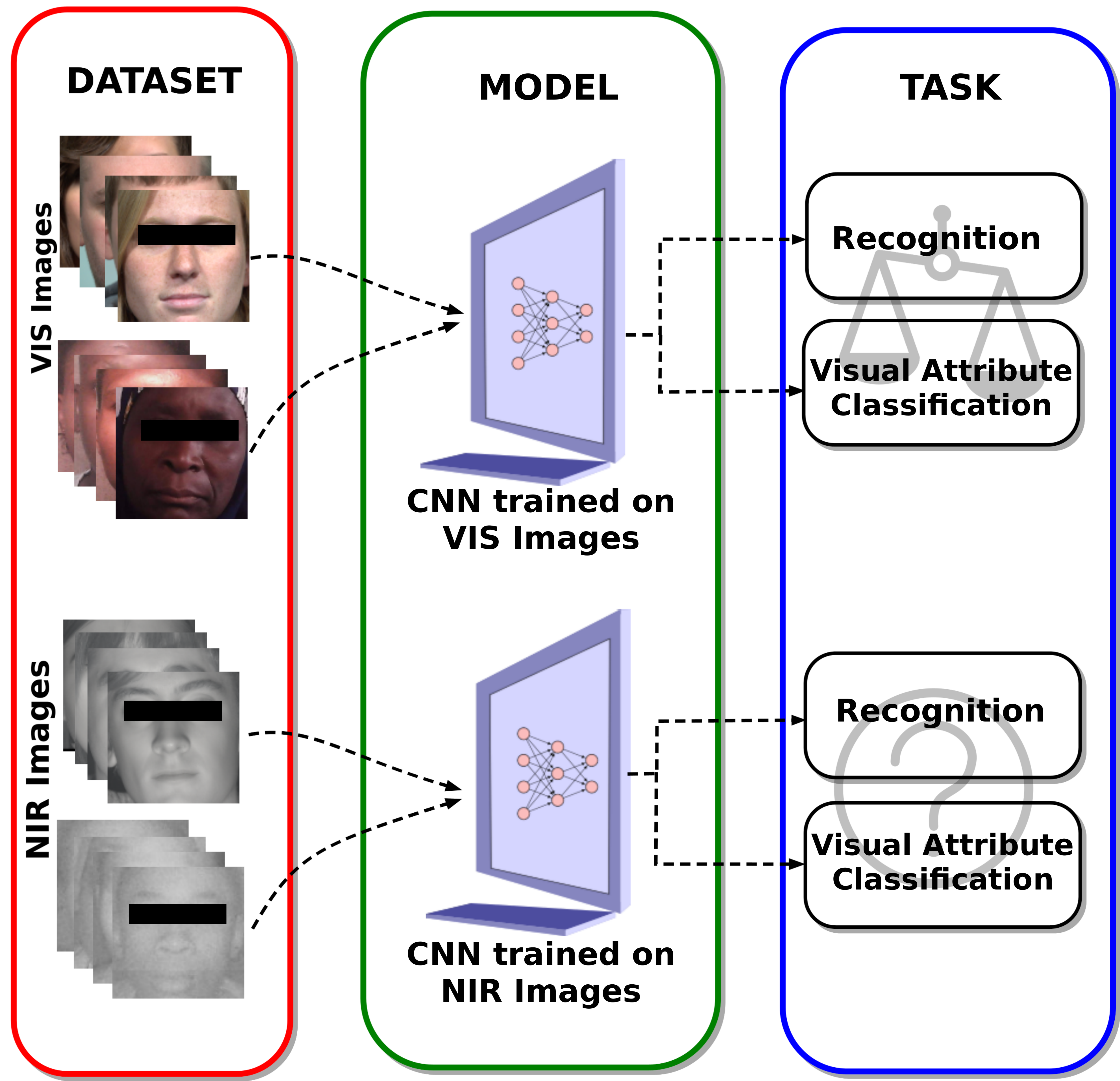}
\caption{\label{fig:schema} \small{Prior studies suggest that VIS-based facial analytics (both recognition and visual attribute classification is biased across race, gender, and age due to factors such as illumination variation, make-up, and skin color. The NIR spectrum is invariant to these factors. Therefore, it would be interesting to examine the bias of facial analytics at the NIR spectrum to obtain more insight into the factors causing  differential performance across demographics.}} 
\end{figure}

Worth mentioning, all these aforementioned studies analyzed facial images at \textbf{visible spectrum} (VIS) ($400 - 800$ nanometers) for bias evaluation.
\textit{These studies suggested that factors such as facial makeup, facial hair, skin color, illumination variation, and facial poses are the main reasons contributing to the bias of facial recognition}~\cite{DBLP:conf/wacv/AlbieroSVZKB20}. 
The NIR spectrum ($800-2500$ nanometers) is invariant to some of these factors (skin color, illumination variation, and facial make up). Therefore, it would be interesting to examine the bias of facial recognition at the NIR spectrum to understand the main cause of the differential performance across demographics (see Figure~\ref{fig:schema}).

\noindent \textbf{Our Contribution.}
The aim of this study is to \emph{investigate the bias of the facial recognition at NIR spectrum}.  To the best of our knowledge, this is the \textbf{first study} of its kind. To this front, the contributions of this work are as follows:

\begin{itemize}
\item An examination of bias of face recognition at near-infrared spectrum~(NIR) for African and Caucasian subjects across gender and race.
\item The curation of NIR based CASIA-Face-Africa~\cite{DBLP:journals/tifs/MuhammadWWZS21} and Notre Dame-NIVL~\cite{DBLP:conf/btas/BernhardBBF15} datasets captured at NIR spectrum for bias evaluation.
\item The use of Explainable AI (XAI) based Gradient Weighted Class Activation Mapping (Grad-CAM)~\cite{DBLP:journals/ijcv/SelvarajuCDVPB20} visualization to understand the distinctive image regions used by the deep learning models across gender and race at NIR spectrum.

\end{itemize}

This paper is organized as follows:  The relevant related work on the VIS spectrum is discussed in Section II. Section III describes the motivation for bias evaluation of face recognition at NIR spectrum. Section IV describes the datasets and models used for training and evaluation. Experimental results and key findings are discussed in Section V. Conclusion and future works are mentioned in Section VI.

\section{Prior Work on Bias of Facial Recognition at VIS}
In this section, we discuss the recent work on the bias of the deep learning-based facial recognition at VIS. 

In a study report by NIST using commercial off-the-shelf (COTS) systems, Grother et al.~\cite{42061} reported that females obtained higher false non-matches at a fixed false match rate~(FMR) than males. Using three COTS systems on mug shot face pictures from the Pinellas County Sheriff's Office, Klare et al.~\cite{DBLP:journals/tifs/KlareBKBJ12} suggested inferior biometrics performance for females, young, and black cohorts.
Krishnapriya et. al~\cite{9001031} conducted studies on MORPH dataset and observed the presence of significant differences in the genuine and impostors score distribution among African American and Caucasian subgroups, and also the same skin-tone image pairs appear more frequently in the impostor distribution. 
Hupont et al.~\cite{8756625} studied the demographic bias of popular public face datasets such as CWF, LFW, VGGFace and VGGFace2 using known CNN models such as FaceNet, SphereFace and ResNet-$50$ and observed higher biometric performance for white males and lower for Asian females. 

Serna et al.~\cite{DBLP:conf/aaai/SernaMFCOR20} reported better biometric performance for White males and the least observed for East Asian females when ResNet-$50$ and VGG-Face models were used on DiveFace dataset. They attributed the performance differential to an imbalanced training set. 
Alberio et al.~\cite{DBLP:journals/tifs/AlbieroZKB22} did cause and effect analysis on the inferior face recognition performance females. 
According to experimental research, the reasons include gendered societal conventions for hairstyle and makeup, as well as morphological differences in face size and shape across gender. 
All the aforementioned studies confirmed strong demographic differences in face recognition performance across gender and race when trained and evaluated on images in visible spectrum.

\section{Motivation behind Examining Bias at NIR Spectrum}


NIR spectrum is invariant to various factors attributed to the differential performance of facial recognition at VIS. The reasons  behind examining bias at NIR spectrum are discussed as follows:
\begin{enumerate}
    \item \textbf{NIR-based face recognition commonly adopted for surveillance:} Most of the facial recognition systems used for surveillance operate at NIR spectrum~\cite{DBLP:journals/csr/FarokhiFS16}. The advantage of the use of near-infrared light for face recognition used for  surveillance are as follows: (1) the near-infrared illuminator is not visible to the human eye and keeps the surveillance operation covert, (2) the captured NIR images are not affected by ambient temperature in comparison to thermal images, and (3) the additional cost involved in using the near-infrared illuminator is relatively low.
 \item \textbf{Illumination invariance:} 
 The use of active near-infrared (NIR) imaging is an effective approach to solving the illumination problem~\cite{DBLP:conf/eccv/SaxenaV16}. NIR imaging is less sensitive to fluctuations in visible light illumination, making it suitable for face recognition under varying illumination conditions or at night (see Figure~\ref{fig:illu-invar}).


\begin{figure} [!h]
    \centering
    \subfloat[VIS Image]{{\includegraphics[width=2cm]{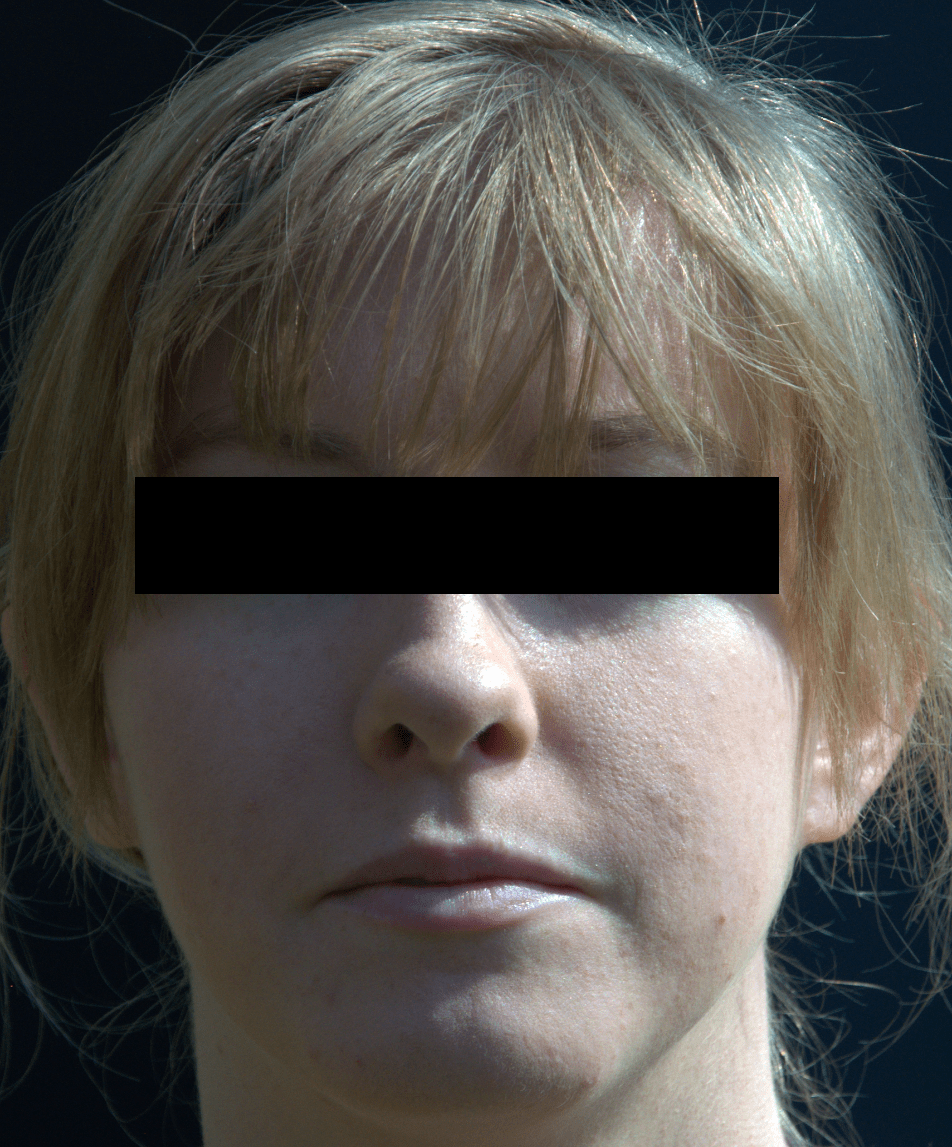} }}
        \hfil
     \subfloat[NIR Image]{{\includegraphics[width=2cm]{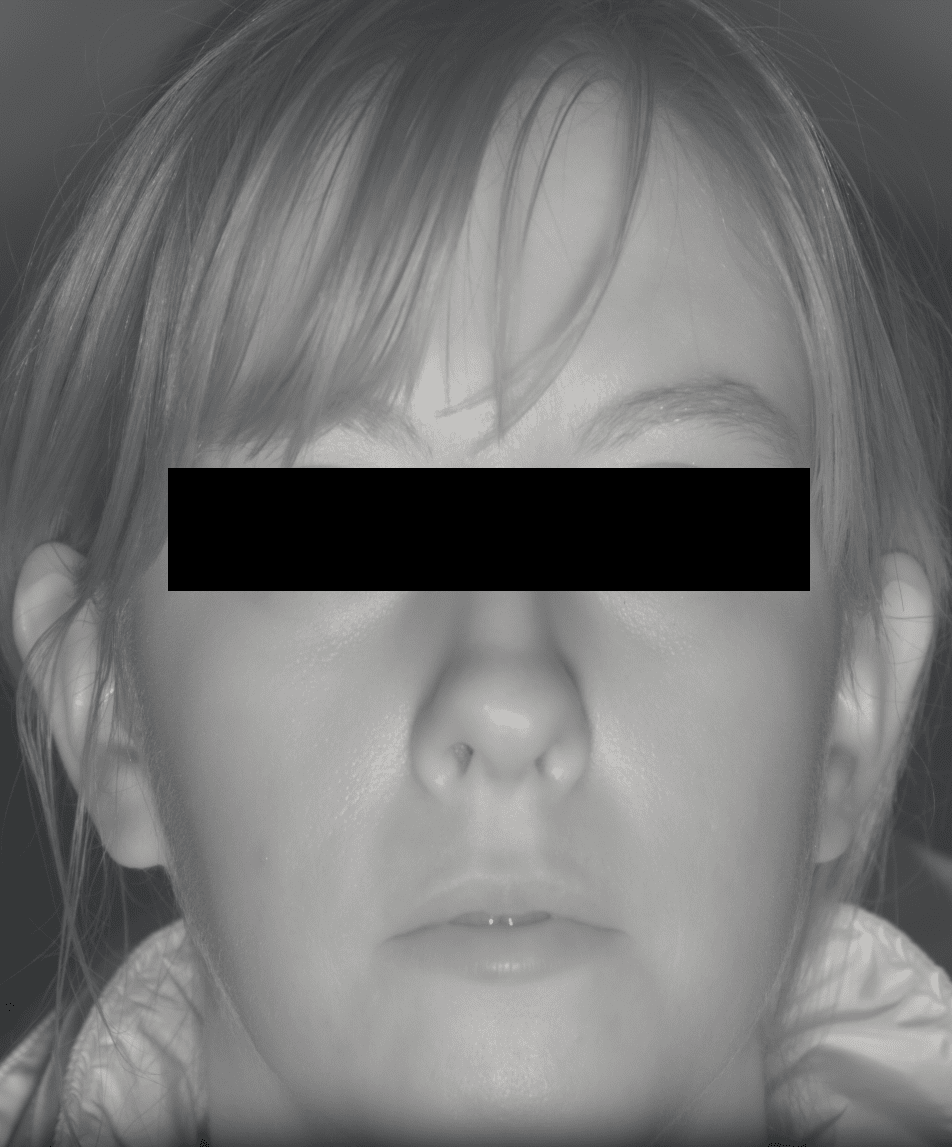} }}
    \caption{Example shows invariance to illumination for NIR facial image compared to VIS facial image.}
    \label{fig:illu-invar}
\end{figure}

\item \textbf{Complementary features captured at NIR:}
As NIR and VIS images are obtained using different sensory modalities, their feature representation differs significantly. For instance, skin texture, skin color, facial shape, and eyelids are captured at VIS whereas at NIR, along with the facial shape and skin texture, iris texture, eyelashes, vein pattern, etc are also captured. Therefore, bias investigation of facial recognition using complementary features learned at the NIR spectrum would facilitate cause and effect analysis. 

\item \textbf{Invariance to facial make-up and skin color:} 
Several studies indicate that there is a high variance in skin reflectance in the VIS spectrum across races attributed to different skin-color types. However, the variance in skin reflectance is almost negligible across race at NIR~\cite{DBLP:conf/mva/KimuraN11}. Further, the impact of facial make-up is minimal at the NIR spectrum (see Figures~\ref{fig:skin-col} and~\ref{fig:make-up}).

\begin{figure} [!h]
    \centering
    \subfloat[VIS Image]{{\includegraphics[width=2cm]{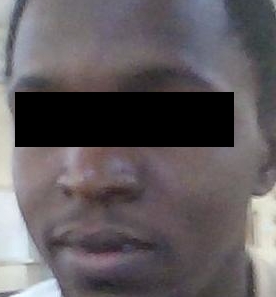} }}
    \hfil
     \subfloat[NIR Image]{{\includegraphics[width=2cm]{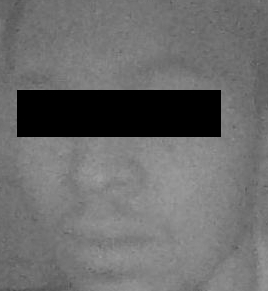} }}
    \caption{Example shows invariance to skin color in an NIR facial image when compared to VIS facial image.}
    \label{fig:skin-col}
\end{figure}
\begin{figure} [!h]
    \centering
    \subfloat[VIS Image]{{\includegraphics[width=2cm]{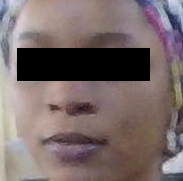} }}
    \hfil
     \subfloat[NIR Image]{{\includegraphics[width=2cm]{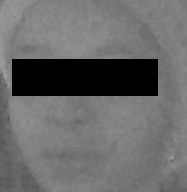} }}
    \caption{Example shows invariance to make-up for a NIR facial image when compared to VIS counterpart facial image.}
    \label{fig:make-up}
\end{figure}

\end{enumerate}

\section{Dataset and Experimental Protocol}
\subsection{Datasets used}
\begin{figure}
    \centering
    \subfloat[VIS WM]{{\includegraphics[width=2cm]{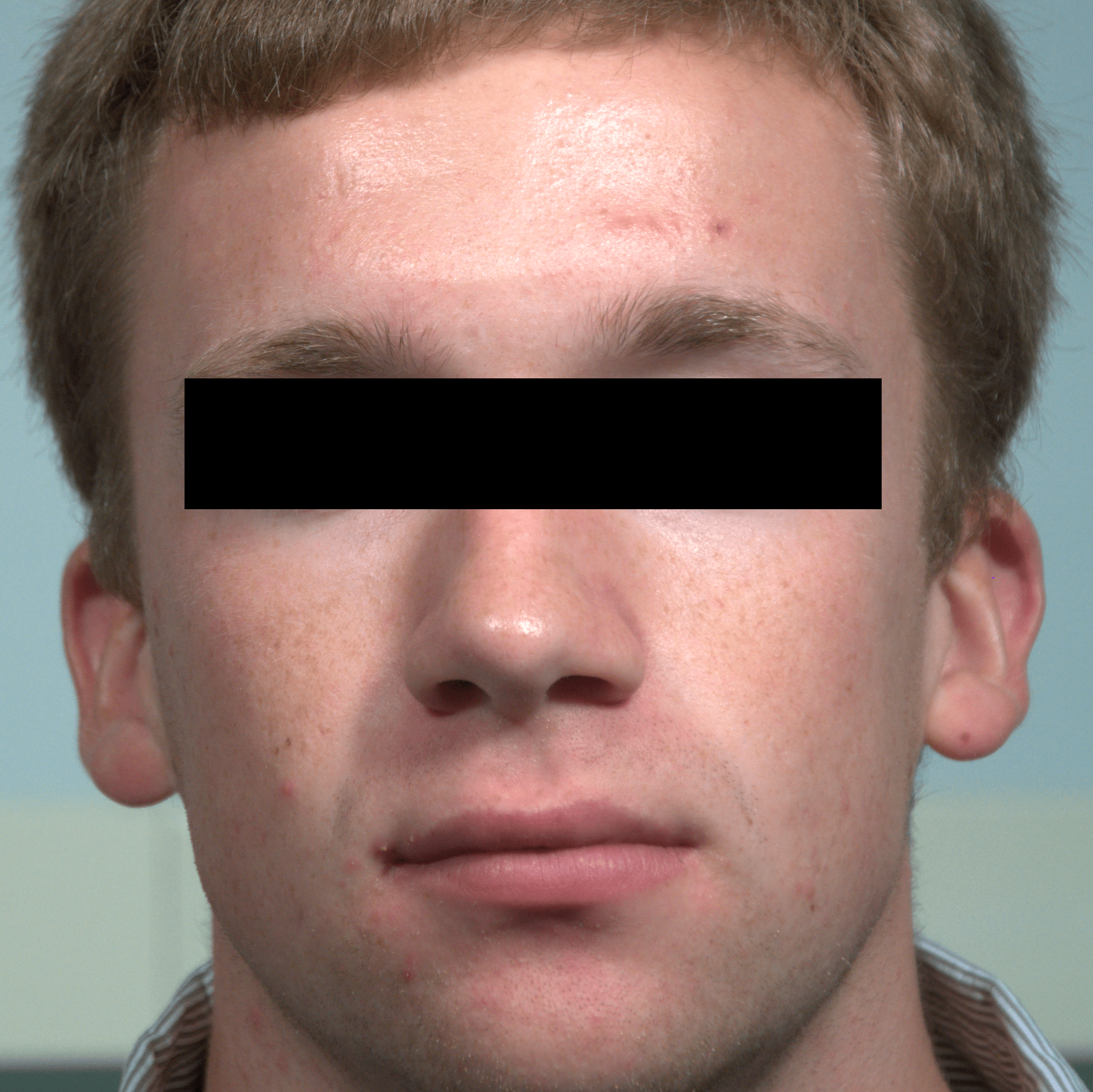} }}
     \subfloat[VIS WF]{{\includegraphics[width=2cm]{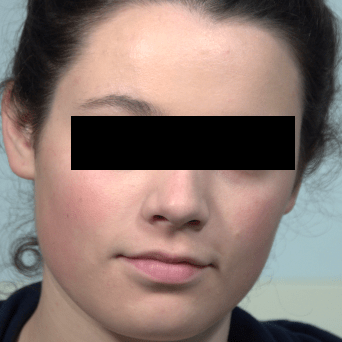} }}
    \hfill
    \subfloat[NIR WM]{{\includegraphics[width=2cm]{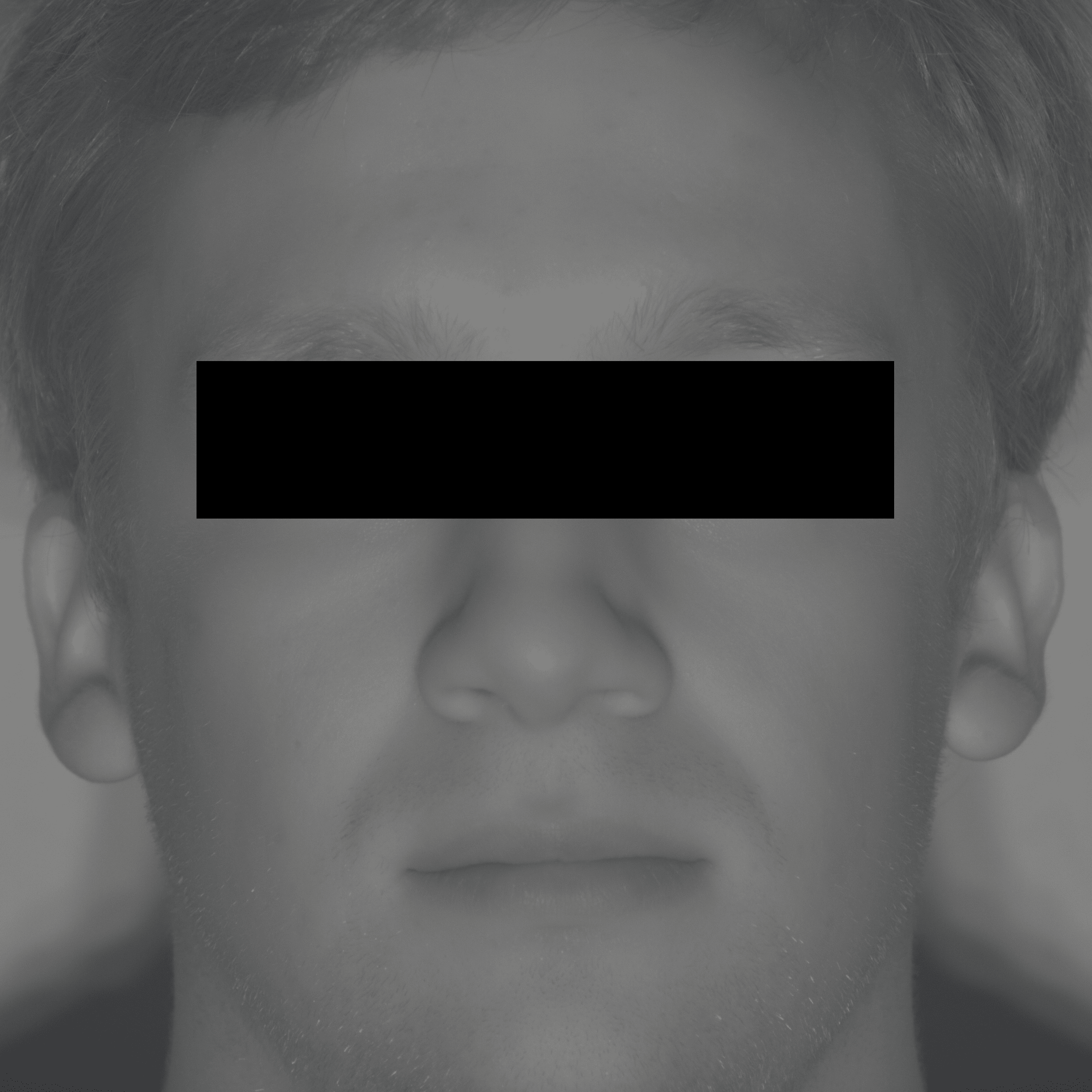} }}
    \subfloat[NIR WF]{{\includegraphics[width=2cm]{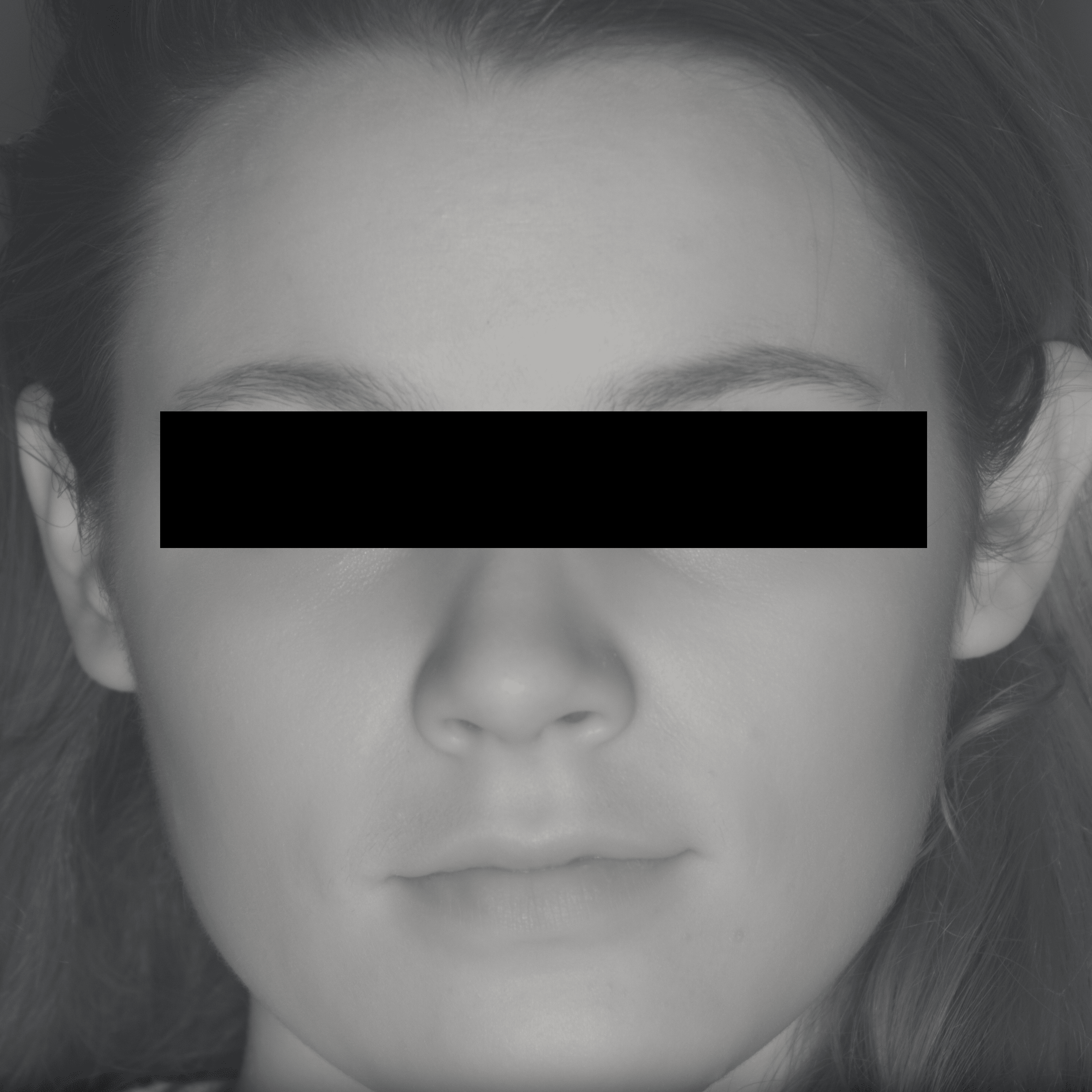} }}
    \caption{Example face images from ND-NIVL Dataset for White (Caucasian) Males~(WM) and Females~(WF) captured at VIS and NIR spectrum.}
    \label{fig:exampleND}
\end{figure}

 The datasets used in this research are described below: 
 \begin{itemize}
 \item \textbf{CASIA-Face-Africa}~\cite{DBLP:journals/tifs/MuhammadWWZS21} is a collection of $38,546$ images of $1,183$ African subjects. Facial images are captured using multispectral cameras in a variety of lighting conditions. The NIR images have a resolution of $983 \times 877$. This dataset was assembled to facilitate research and development on the racial bias of the face recognition system for dark-skinned people. Figure~\ref{fig:exampleAfrica} shows the example face images from the CASIA-Africa dataset for Black Males~(BM) and Black Females~(BF) captured at VIS and NIR spectrum, respectively.
 
 

\begin{figure}
    \centering
    \subfloat[VIS BM]{{\includegraphics[width=2cm]{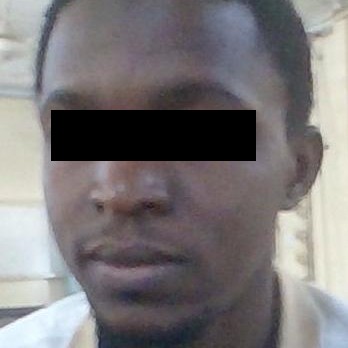} }}\subfloat[VIS BF]{{\includegraphics[width=2cm]{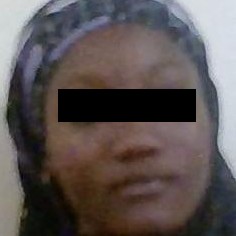} }}\hfill \subfloat[NIR BM]{{\includegraphics[width=2cm]{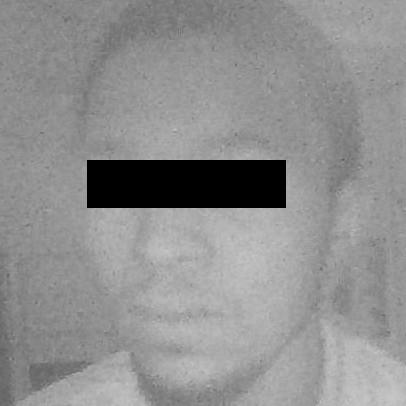} }}\subfloat[NIR BF]{{\includegraphics[width=2cm]{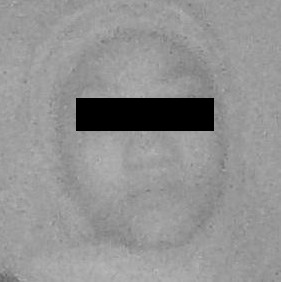} }}
    \caption{Example facial images from CASIA-Face-Africa Dataset for Black Males~(BM) and Females~(BF) captured at NIR and VIS spectrum.}
    \label{fig:exampleAfrica}
\end{figure}
 
 \item \textbf{Notre Dame Near-Infrared and Visible-Light (ND-NIVL)}~\cite{DBLP:conf/btas/BernhardBBF15} is the largest available NIR/VIS  dataset with $24,605$ images of $574$ subjects.  All images were acquired under normal indoor lighting with frontal pose and neutral facial expression. The NIR images have a resolution of $4770\times3177$. This makes ND-NIVL the largest database of high-resolution NIR images. Figure~\ref{fig:exampleND} shows the example face images from Notre Dame dataset for (Caucasian) White Males~(WM) and White Females~(WF) captured at VIS and NIR spectrum.
 
 \end{itemize}
Worth mentioning, that these two datasets are the only available facial image datasets containing male and female subjects from Caucasian and Black subjects captured at the NIR spectrum. 
\begin{table}[hbt!]
\centering
\caption{Gender and Race Balanced Dataset Distribution for Facial Recognition.}
\resizebox{\linewidth}{!}{%
\label{tab:_datafr}
\begin{tabular}{|c|cccccc|}
\hline
 & \multicolumn{6}{c|}{\textbf{\#Images}} \\ \hline
\textbf{} &
  \multicolumn{1}{c|}{\textbf{\%}} &
  \multicolumn{1}{c|}{\textbf{Total}} &
  \multicolumn{1}{c|}{\textbf{WM}} &
  \multicolumn{1}{c|}{\textbf{WF}} &
  \multicolumn{1}{c|}{\textbf{BM}} &
  \textbf{BF} \\ \hline
\textbf{Train} &
  \multicolumn{1}{c|}{68} &
  \multicolumn{1}{c|}{5949} &
  \multicolumn{1}{c|}{1514} &
  \multicolumn{1}{c|}{1509} &
  \multicolumn{1}{c|}{1487} &
  1439 \\ \hline
\textbf{Valid/Test} &
  \multicolumn{1}{c|}{22/10} &
  \multicolumn{1}{c|}{1901/883} &
  \multicolumn{1}{c|}{495/221} &
  \multicolumn{1}{c|}{496/222} &
  \multicolumn{1}{c|}{467/222} &
  443/218 \\ \hline
\end{tabular}%
}
\end{table}
\\
\noindent \textbf{Curation:}  Unlike CASIA-Africa, the NotreDame-NIR dataset contained subjects from other races as well apart from Caucasians. Both the datasets were curated to develop a gender and race-balanced training set and to ensure only white subjects were present in the dataset. The term Caucasian and White are used interchangeably and denote the same thing.

As the ND-NIR dataset had a very diverse set of subjects, each of the $574$ subjects was manually categorized into racial groups to identify and isolate only white subjects. The CASIA-Face-Africa dataset was also manually reviewed, but it contained all-black subjects.

The CASIA-Face-Africa dataset is annotated with gender information for each of the subjects, which was spot-checked for accuracy. However, the Notre Dame-NIR dataset is not annotated with gender labels. Therefore, gender labels are manually assigned based on the visual inspection by the graduate research assistant. 
Next, Dlib, a toolkit for making real-world machine learning and data analysis, majorly used for face detection and facial landmark extraction, was used to extract the faces from all subjects in both datasets. The number of images for each subject varied in both datasets. For each of the subjects, $8$ to $12$ images were randomly selected for the training and testing set.




Table~\ref{tab:_datafr} displays the distribution of racial and gender data for Black and White Males and Females (produced by CASIA-Africa and Notre Dame) utilized for bias study of facial recognition at the NIR spectrum. Since the majority of the research in this subject adheres to the idea that "gender" as binary, we also adhere to it for the sake of fair comparison. We do not intend to belittle people who disagree with this view in the course of conducting this study.

For the facial recognition experiments, a train/test split of 90/10 was used in a subject-independent manner. Among $90\%$ of subjects present in the training dataset, $80\%$ of each subject's images were used for the training set and the other $20\%$ were used for the validation set. 

\subsection{Experimental protocol}

For face recognition, three deep learning models were trained as follows: ResNet-50~\cite{DBLP:conf/cvpr/HeZRS16},  LightCNN~\cite{9320939}, and DenseNet-$121$~\cite{DBLP:conf/cvpr/HuangLMW17}. DenseNet-$121$ was used in conjunction with the ArcFace loss function~\cite{DBLP:conf/cvpr/DengGXZ19}. 
The pre-trained version of these models, on Imagenet weights, were fine-tuned by adding a fully-connected layer of size $512$ followed by an output layer. 
The fine-tuning was performed using an empirically found batch size of $32$ along with Stochastic Gradient Descent (SGD) optimizer with the Cosine Annealing with a warm restart learning scheduler. The models were trained using an early stopping mechanism.

\section{Experimental results}
Table~\ref{tab:frsum} summarizes the results for all models for facial recognition. LightCNN obtained the best performance metrics. ArcFace obtained the worst performance with a ROC-AUC score under $0.90$. 

For each model, the distribution of genuine and imposter scores have been plotted in Figures~\ref{fig:score-distribution-resnet},  \ref{fig:score-distribution-lcnn} and \ref{fig:score-distribution-arcface}. Black and White subjects were divided into separate graphs with male and female subjects being plotted with their curves in the same graph.
The genuine score distribution for \textit{black subjects} tended to be more widely distributed than for white subjects. The distance between the genuine and imposter distribution is described by the d-prime score. A higher d-prime indicates that the models were better able to perform the facial recognition task, or high true matches and low false matches.  Male and female d-prime scores for each race were quite similar, with differences between them ranging between $0.0$ and $1.0$. This would imply that there is an insignificant difference between the performance of males and females. 

The males outperformed females for every model for \textit{white subjects}; however, there is no clear trend of males or females performing better for black subjects. When we compare the d-prime scores of black and white individuals, we can observe that each model is biased toward white subjects, because the d-prime scores for both male and female subjects were roughly $1.0$ to $2.0$ times higher than that of black subjects. This may be because of the difference in sensor quality used for imaging. 

Across races, the genuine score distributions between males and females remain relatively similar across each race and model, however, white subjects have slightly higher overall scores. 
The distribution of genuine scores did vary slightly based on race and gender, however, the overall distributions were very similar across gender and race.

The imposter scores and distributions were consistent across both gender and race. The distributions for males and females lie almost entirely on top of one another for each race. Each of the imposter distributions was centered around very similar thresholds. These scores centered around $0.3$ to $0.5$ for all models apart from the worst model, ArcFace, in Figure~\ref{fig:score-distribution-arcface} which centered around $0.2$ to $0.3$, which indicated that ArcFace had lower imposter scores overall. LightCNN had the lowest imposter scores while also maintaining high genuine scores, unlike ArcFace. 
\textbf{In summary, neither gender nor race appear to have a significant impact on the imposter scores of any of the models.}


\begin{strip}
\centering
\includegraphics[scale=0.45]{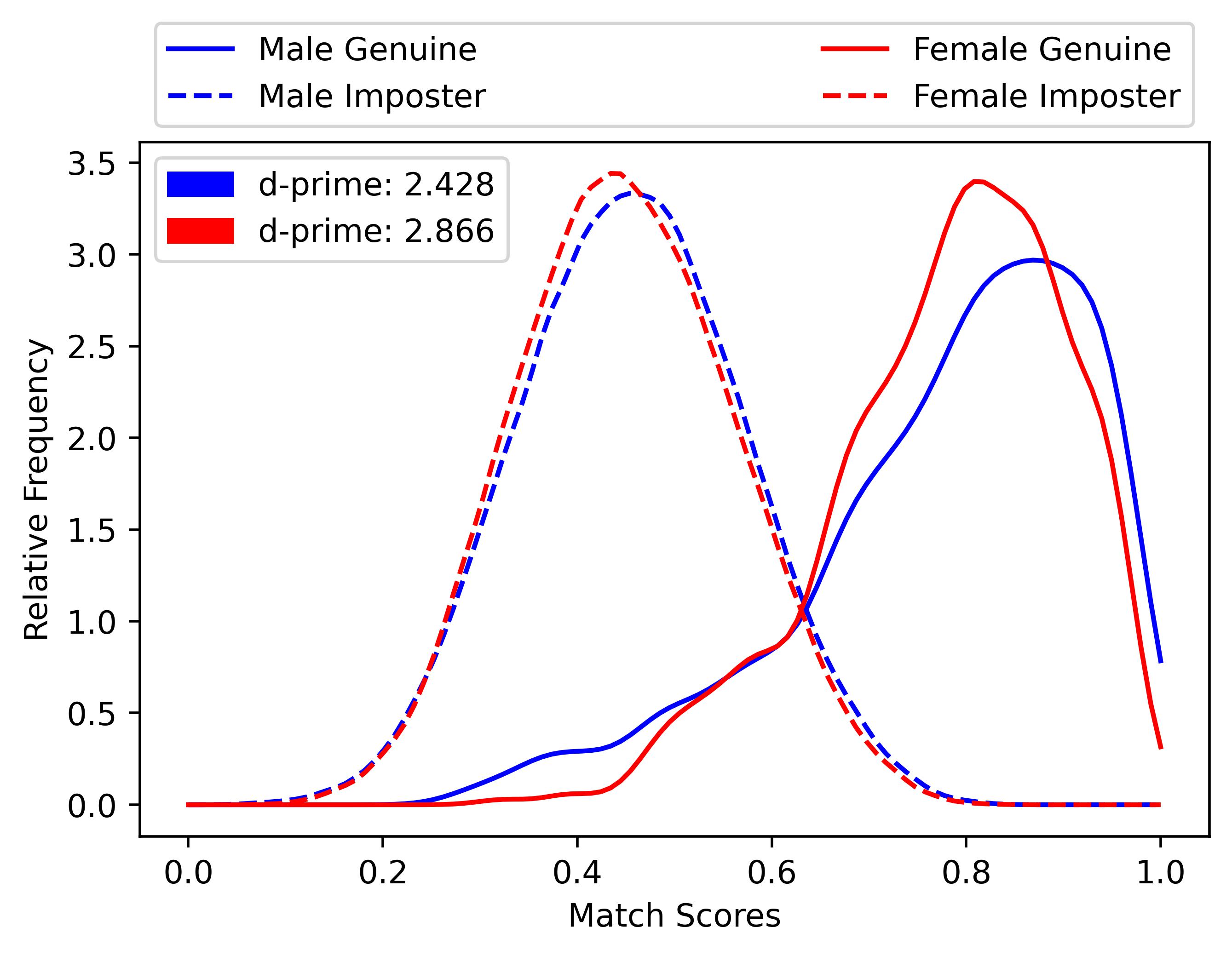}\hfil
\includegraphics[scale=0.45]{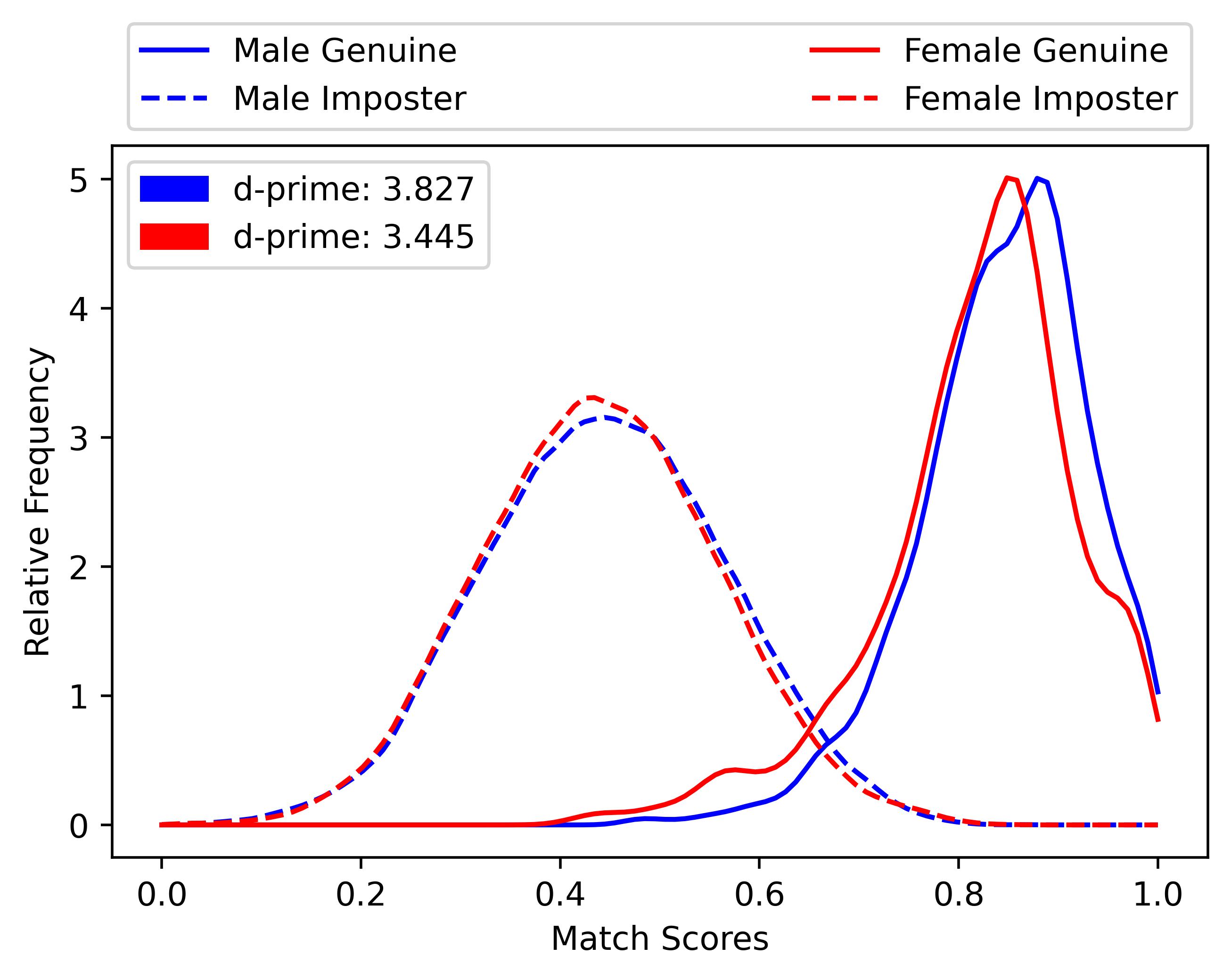}\hfil
\vspace{-2ex}
\captionof{figure}{Score distributions for black (left) and white (right) subjects for ResNet-50.}
\label{fig:score-distribution-resnet}
\includegraphics[scale=0.45]{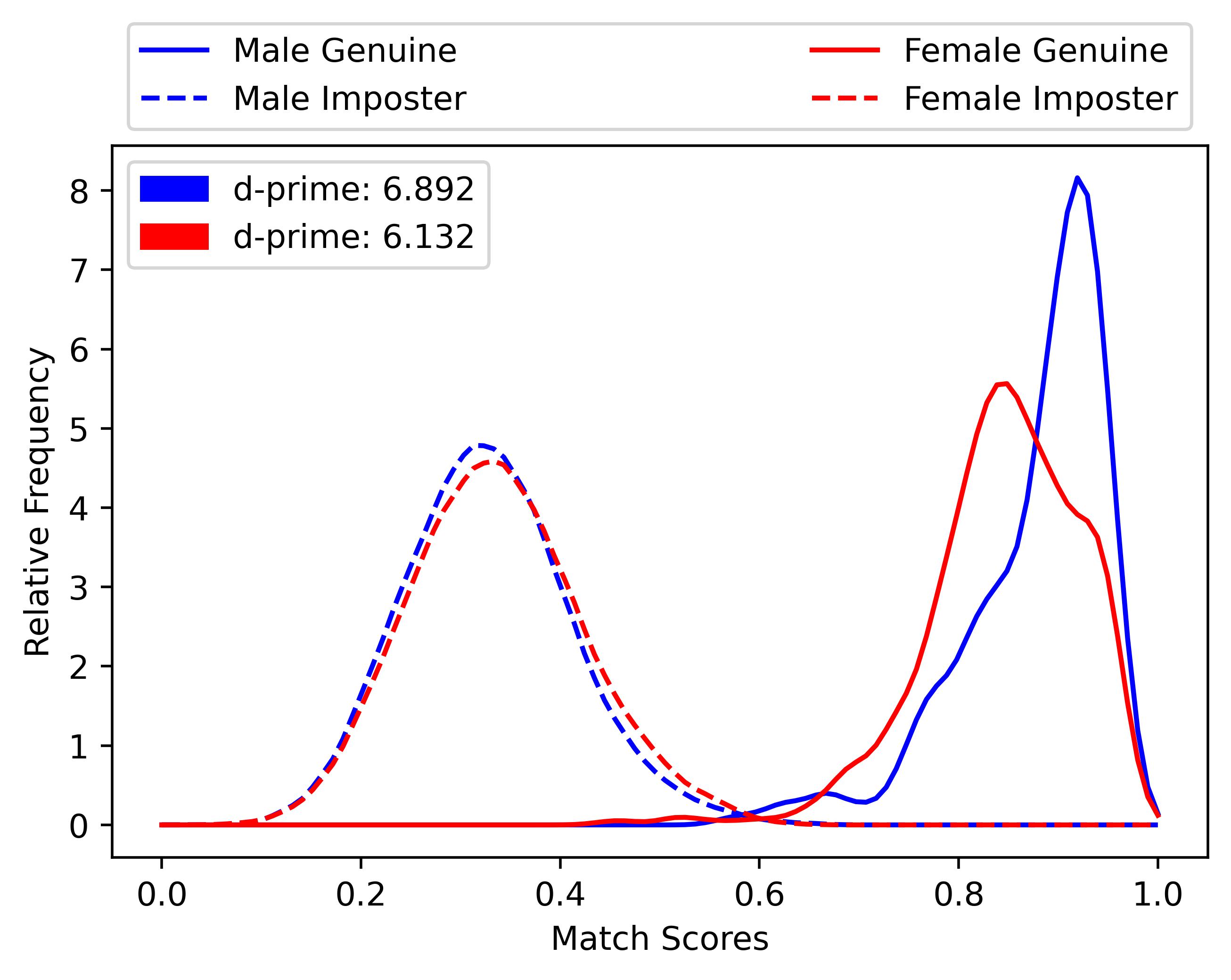}\hfil
\includegraphics[scale=0.45]{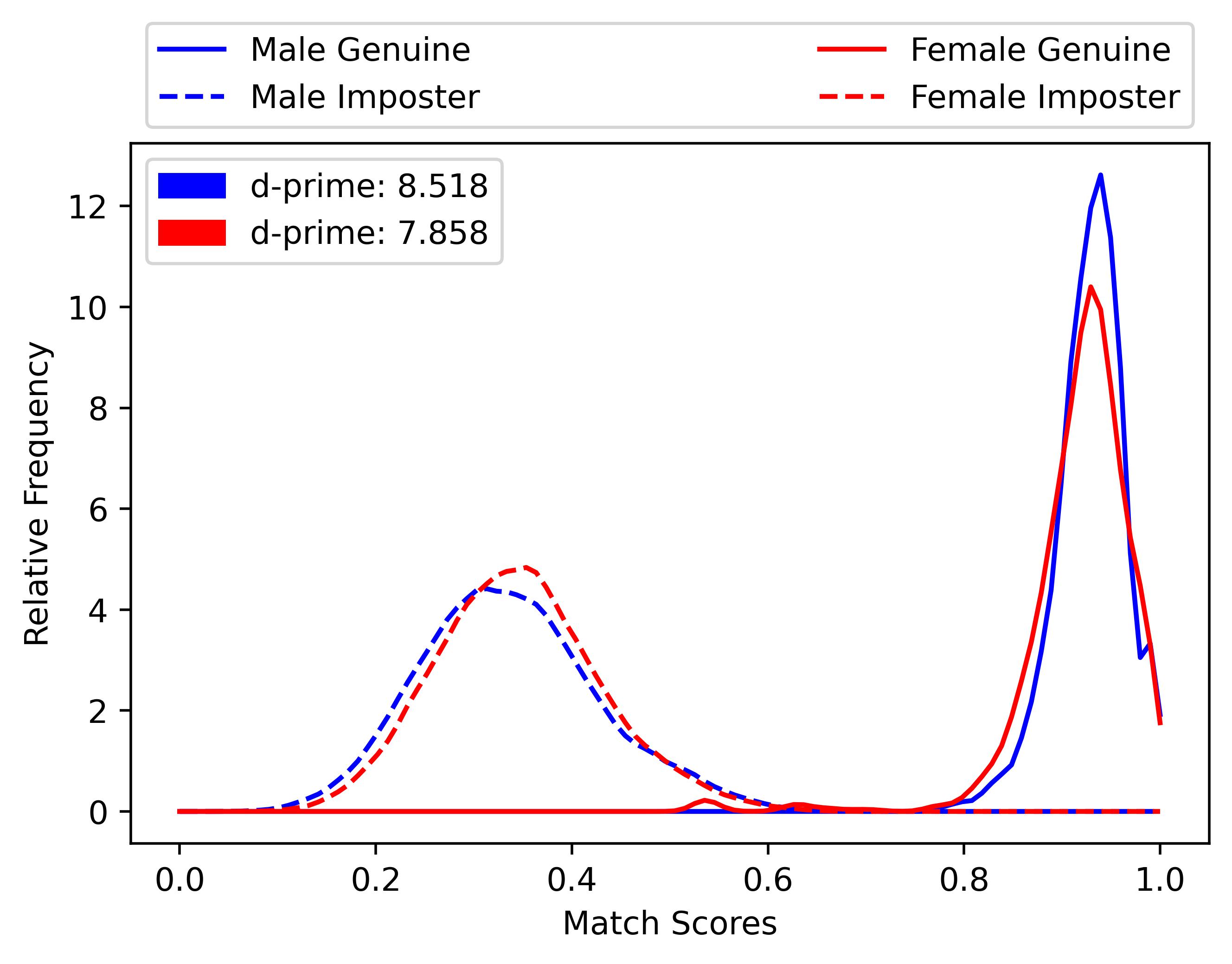}\hfil
\vspace{-2ex}
\captionof{figure}{Score distributions for black (left) and white (right) subjects for LightCNN.}
\label{fig:score-distribution-lcnn}


\includegraphics[scale=0.45]{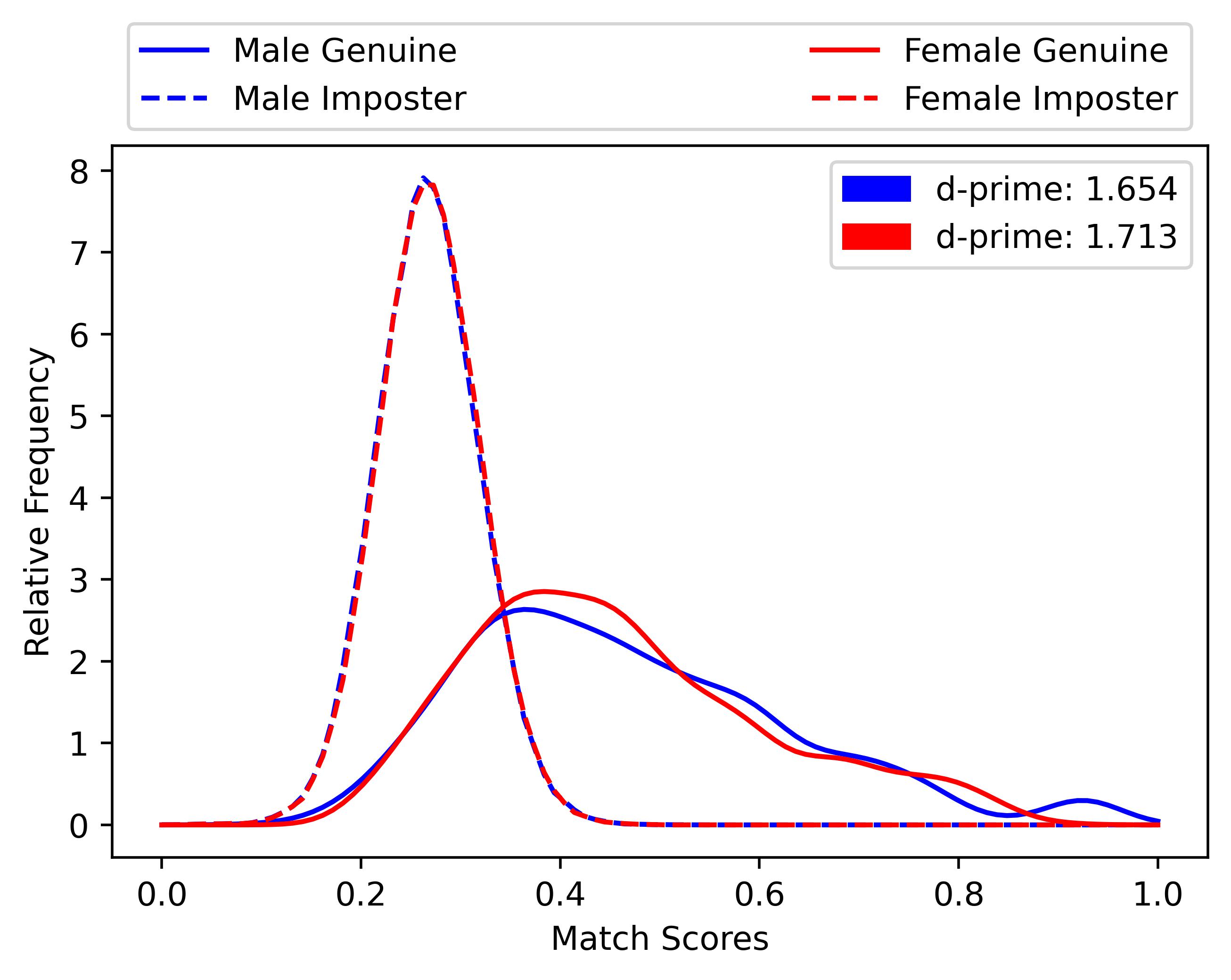}\hfil
\includegraphics[scale=0.45]{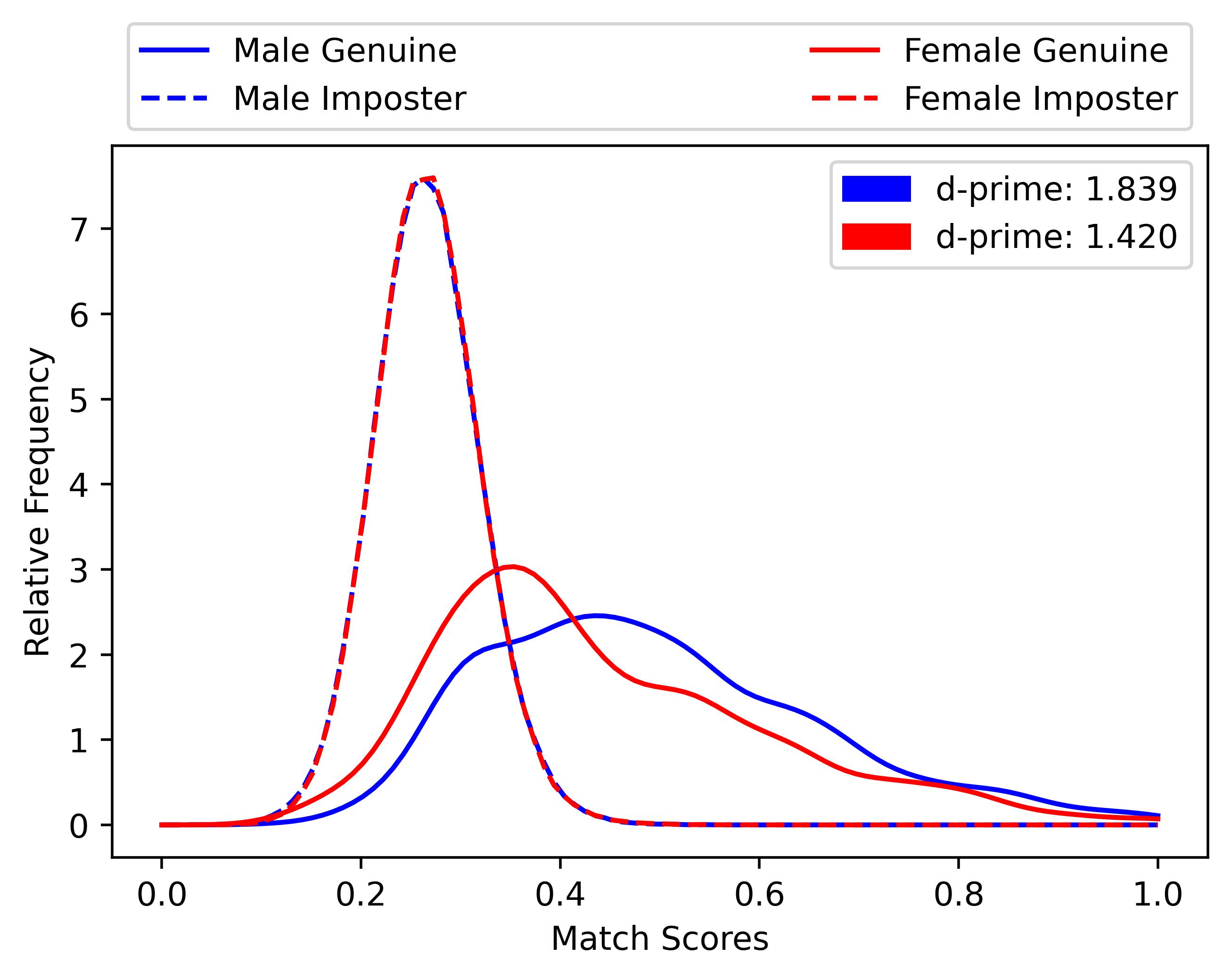}\hfil
\vspace{-2ex}
\captionof{figure}{Score distributions for black (left) and white (right) subjects for ArcFace.}
\label{fig:score-distribution-arcface}
\end{strip}
\vspace{-5mm}

\begin{table*}[hbt!]
\centering
\caption{Facial Recognition Model Performance Summary}
\label{tab:frsum}
\resizebox{0.5\linewidth}{!}{%
\begin{tabular}{|c|c|c|c|}
\hline
\textbf{Model}          & \textbf{ResNet-50} & \textbf{LightCNN} & \textbf{ArcFace} \\ \hline
\textbf{ROC-AUC}        & 0.974              & 0.999             & 0.899            \\ \hline
\textbf{EER}            & 7.7\%              & 0.5\%             & 17.8\%           \\ \hline
\textbf{TPR@0.001\%FPR} & 34\%               & 98\%              & 23\%             \\ \hline
\textbf{TPR@0.010\%FPR} & 51\%               & 98\%              & 34\%             \\ \hline
\textbf{TPR@0.100\%FPR} & 66\%               & 99\%              & 46\%             \\ \hline
\textbf{TPR@0.500\%FPR} & 76\%               & 99\%              & 55\%             \\ \hline
\end{tabular}%
}
\end{table*}

\begin{strip}
\centering
\includegraphics[scale=0.45]{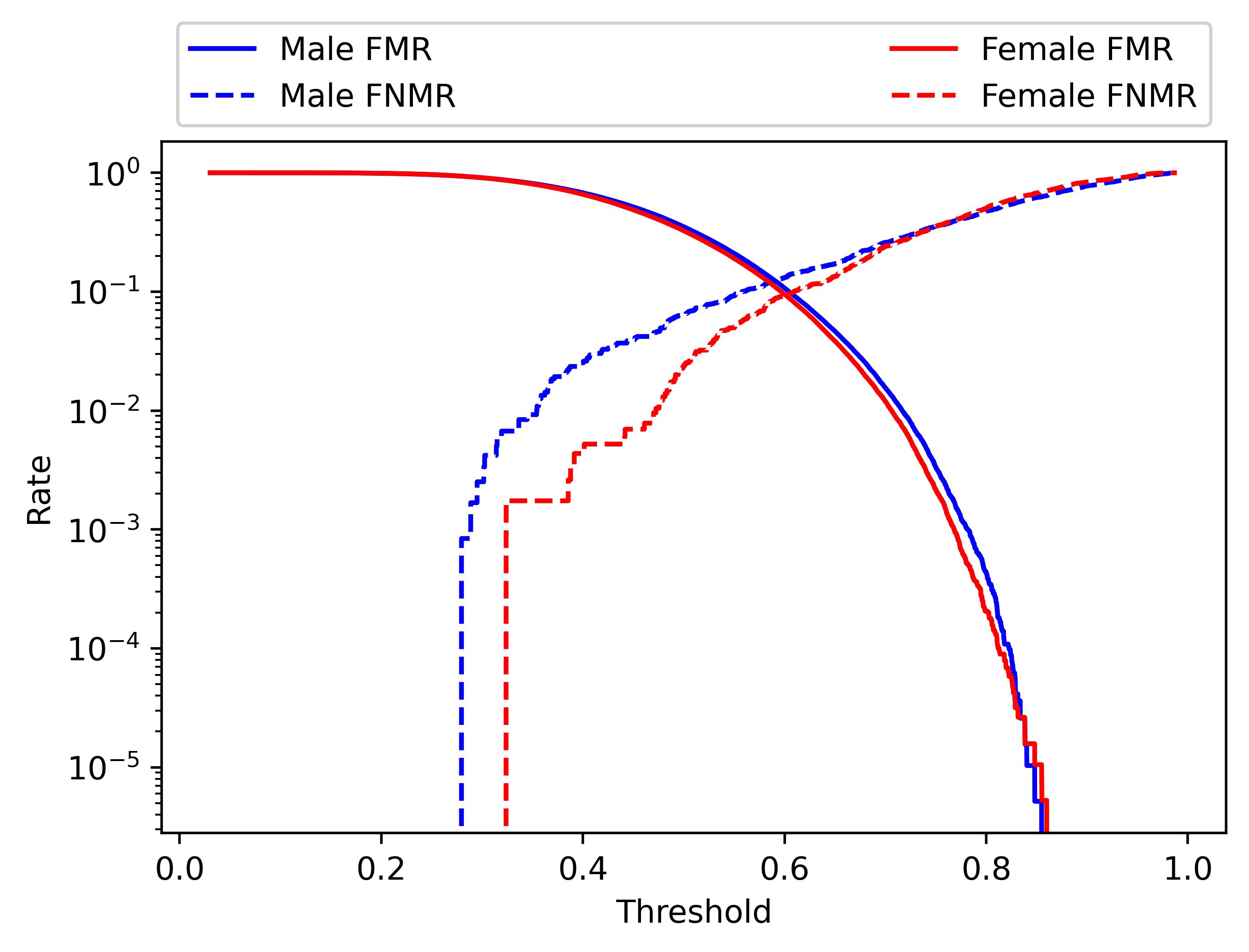}\hfil
\includegraphics[scale=0.45]{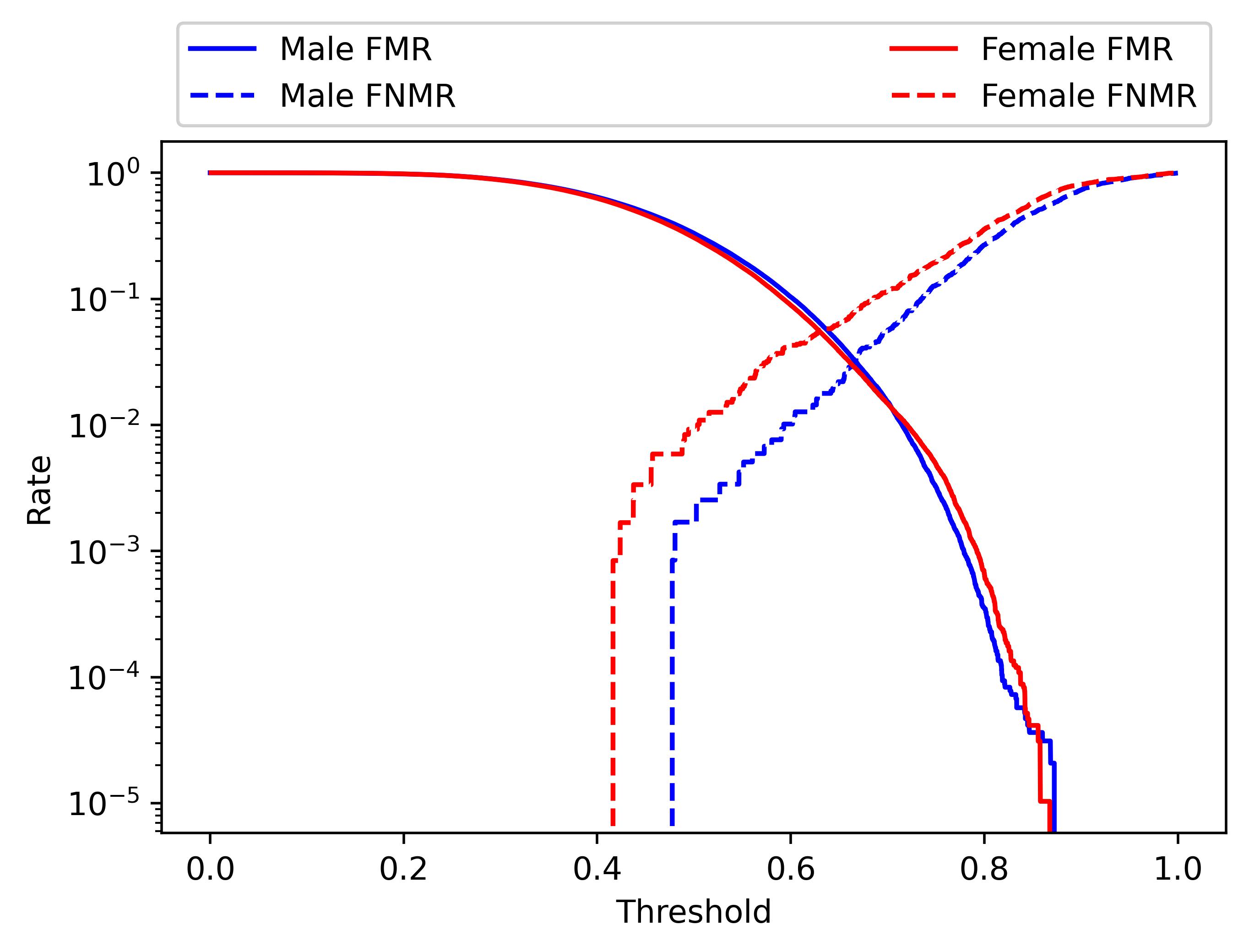}\hfil
\vspace{-2ex}
\captionof{figure}{FMR and FNMR Curves for black (left) and white (right) subjects for ResNet-50}
\label{fig:fmr-fnmr-resnet}


\includegraphics[scale=0.45]{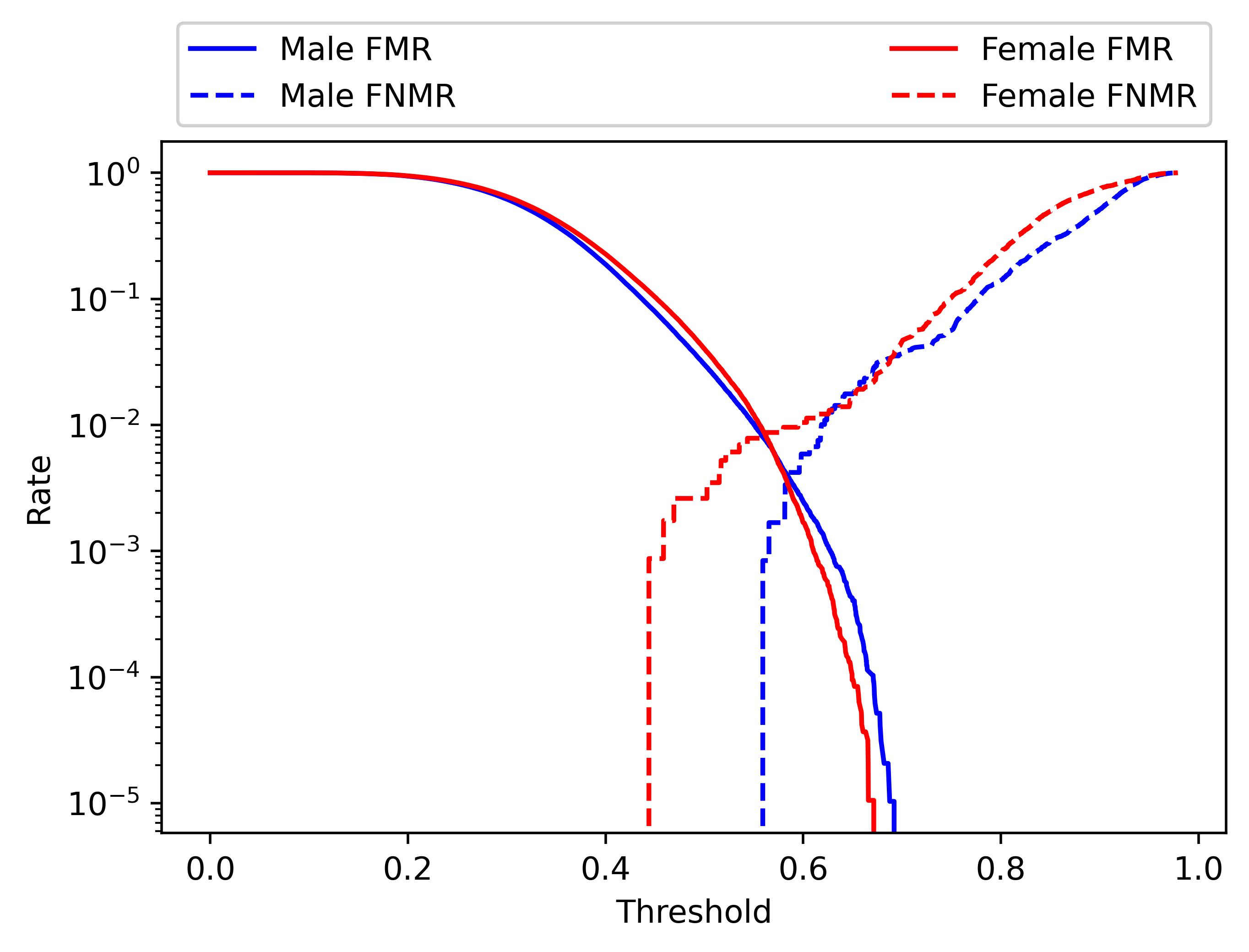}\hfil
\includegraphics[scale=0.45]{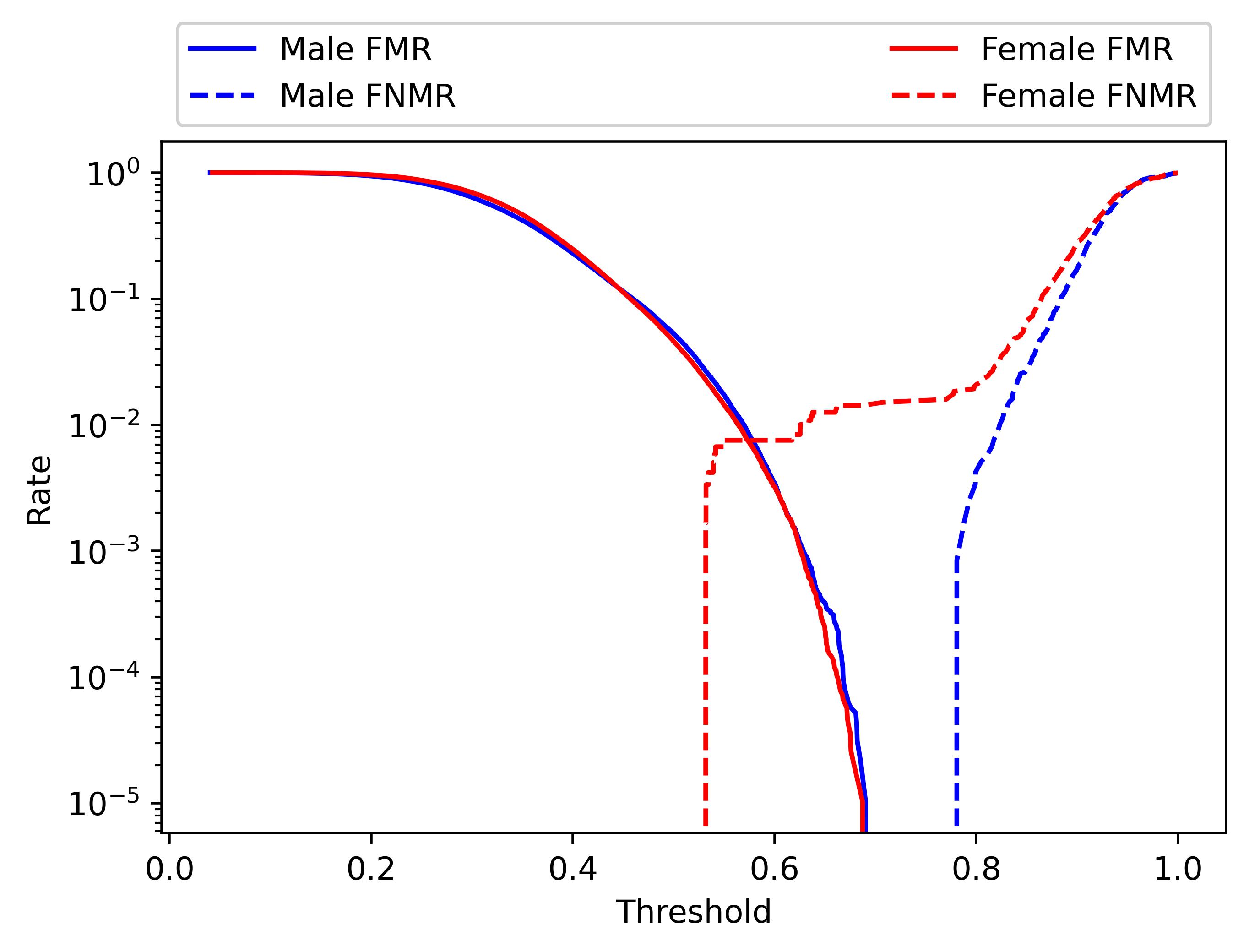}\hfil
\vspace{-2ex}
\captionof{figure}{FMR and FNMR Curves for black (left) and white (right) subjects for LightCNN}
\label{fig:fmr-fnmr-lightcnn}

\includegraphics[scale=0.45]{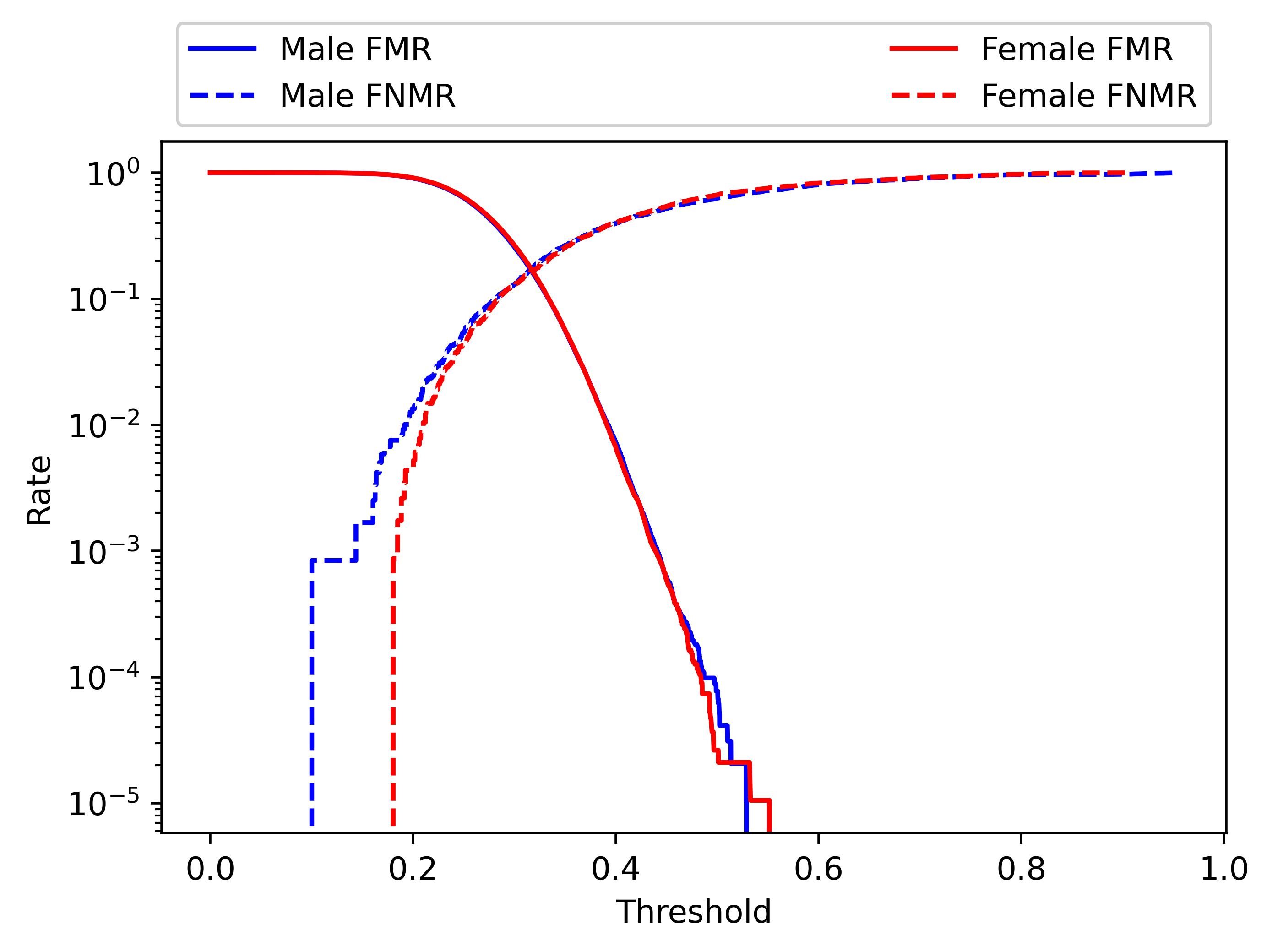}\hfil
\includegraphics[scale=0.45]{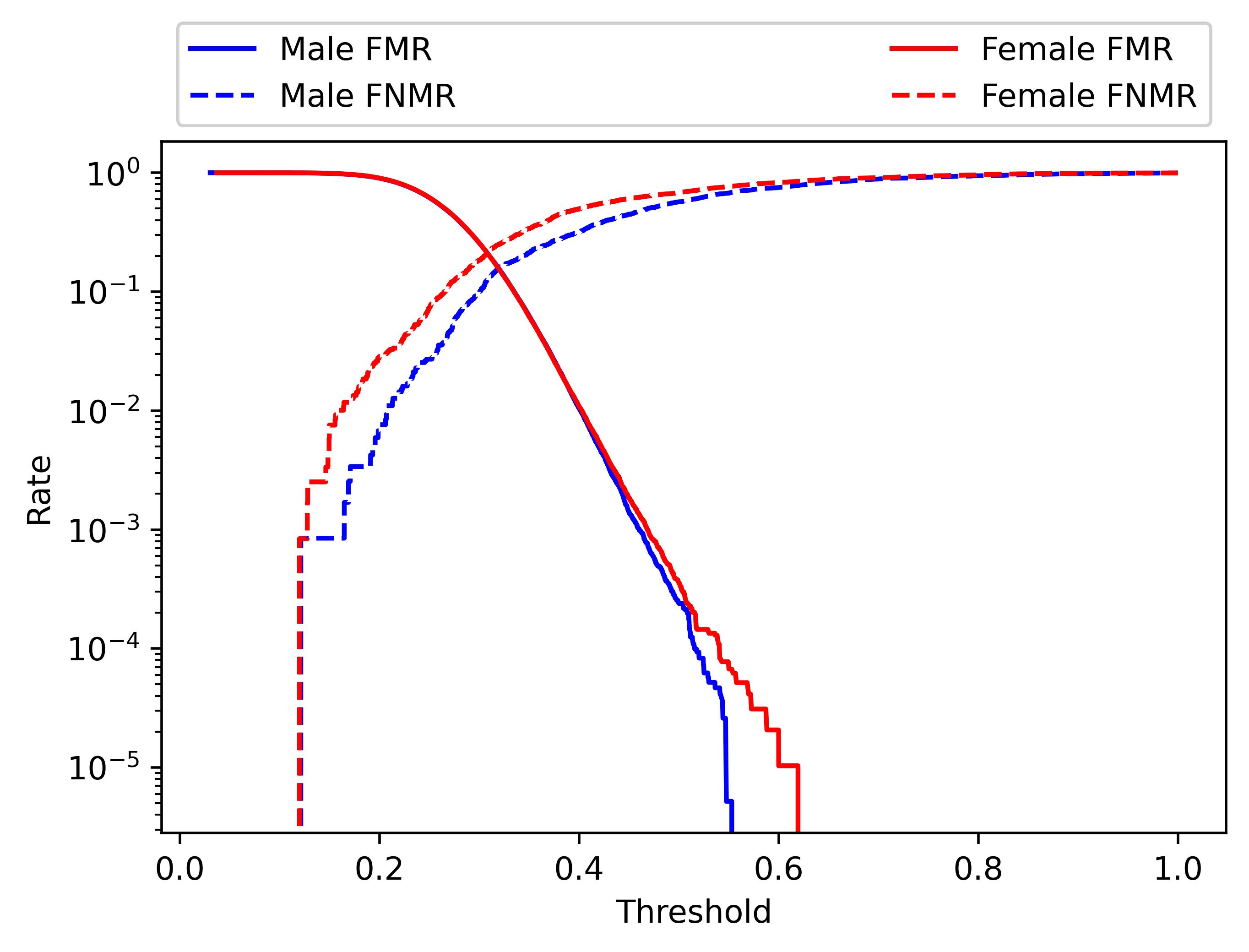}\hfil
\vspace{-2ex}
\captionof{figure}{FMR and FNMR Curves for black (left) and white (right) subjects for ArcFace.}
\label{fig:fmr-fnmr-arcface}
\end{strip}

\vspace{0.5mm}
False Match Rate (FMR)- False Non Match Rate (FNMR) curves (Figures~\ref{fig:fmr-fnmr-resnet}, \ref{fig:fmr-fnmr-lightcnn} and \ref{fig:fmr-fnmr-arcface}) were plotted for each model, segregated by race, with both male and female plots on the same graph, just as the impostor and genuine distributions. The first thing to note in these graphs is,  how closely the FMR curves for males and females resemble each other. The FNMR curves, on the other hand, will tend to diverge at lower thresholds. At a threshold of roughly $0.6$, the FMR and FNMR curves tends to intersect. The curves in ArcFace, on the other hand, crossed closer to $0.35$. For black individuals, females had a higher FNMR than males, while the inverse was true for white subjects.

In each of the Figures~\ref{fig:fmr-fnmr-resnet}, \ref{fig:fmr-fnmr-lightcnn} and \ref{fig:fmr-fnmr-arcface}, we see that the male and female FMR curves lie on top of each other indicating males and females are performing the same in their FMR curves. Furthermore, for each model, the FMR curves for black and white subjects are comparable, ending at nearly identical thresholds. This would imply that similar performance is observed across all races. 

\begin{figure}
\centering
\includegraphics[scale=0.45]{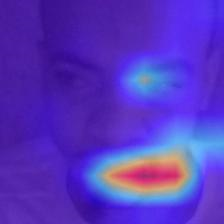}
\includegraphics[scale=0.45]{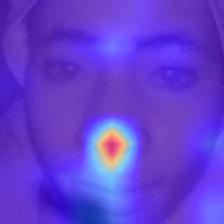}\hfil
\captionof{figure}{Grad-CAM visualization for black male and female subject using ResNet-50.}
\label{fig:cambm}
\end{figure}

\begin{figure}
\centering
\includegraphics[scale=0.45]{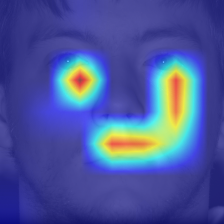}
\includegraphics[scale=0.45]{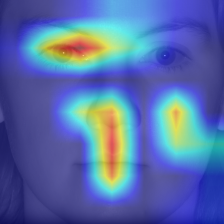}\hfil
\captionof{figure}{Grad-CAM visualization for white male and female subject using ResNet-50.}
\label{fig:camwm}
\end{figure}

Furthermore, we used \textit{Gradient-weighted Class Activation Mapping (GRAD-CAM)} as a technique for XAI to better understand the unique image regions used by CNN models in facial recognition across gender and race at NIR. GRAD-CAM generates a coarse localization map from any target concept's gradients, highlighting different image regions that can be utilized to make a decision/prediction. We used fine-tuned ResNet-$50$ model for this task. The activation from the final fully connected feature layer is considered. The highly activated region is shown by the red zone on the map, followed by green and blue zones.

For black subjects (see~Figure~\ref{fig:cambm}), \textbf{the lip and nose regions} tend to be the most activated. Similarly, \textbf{the nose, eye and the lips regions} were the most activated areas for white subjects (see~Figure~\ref{fig:camwm}). Facial regions with distinct features like the chin, cheeks, and forehead were typically ignored. \textit{Overall, males and females obtained similar activation areas for each race. The distinct image regions were used by the CNN across race.} However, the insignificant differences in performance are observed across gender and race at NIR spectrum. 
\textbf{In summary, current experimental results demonstrates insignificant difference in the performance of the facial recognition across gender and race at the NIR spectrum.}

 
\section{Conclusion and Future Works}

This paper aimed to study the bias of facial recognition across gender and race at NIR spectrum. To facilitate this analysis, Notre-Dame and CASIA-Africa datasets were curated to create a gender and race-balanced subset consisting of African and Caucasian male and female subjects. Based on our experimental results, males and females performed equitably (with insignificant difference) in face recognition for both races. This is contrary to studies at VIS where males consistently outperformed females for face recognition. Across races, white subjects have slightly better performance than black subjects on an average. However, this difference is majorly attributed to the difference in the sensor quality.
This study suggest the merit of the NIR spectrum in reducing demographic bias of facial recognition. There are limited NIR facial datasets annotated with demographic labels. 
This work echoes the importance of benchmarking NIR face datasets for diverse population sub-groups in understanding and validating the fairness of face recognition at NIR spectrum.
The scope of this paper is limited to evaluating bias of facial recognition at NIR spectrum across gender and race. As a part of future work, comparative analysis on the fairness of facial recognition at VIS and NIR spectrum will be performed across gender and race. Further, multimodal systems based on both VIS and NIR images should be investigated for bias mitigation.


\section{Acknowledgements}
This work is supported in part from National Science Foundation (NSF) award no. $2129173$. This work was done as a part of Master Thesis by Brian Neas.
The authors would like to express their thanks to the reviewers for providing numerous suggestions that resulted in improvement in the content and presentation of the paper.
\small
\bibliographystyle{plain}
\bibliography{biblio}

\begin{thebibliography}{10}

\bibitem{DBLP:conf/wacv/AlbieroSVZKB20}
V{\'{\i}}tor Albiero, Krishnapriya~K. S, Kushal Vangara, Kai Zhang, Michael~C.
  King, and Kevin~W. Bowyer.
\newblock Analysis of gender inequality in face recognition accuracy.
\newblock In {\em {IEEE} {WACV} Workshops}, pages 81--89. {IEEE}, 2020.

\bibitem{DBLP:journals/tifs/AlbieroZKB22}
V{\'{\i}}tor Albiero, Kai Zhang, Michael~C. King, and Kevin~W. Bowyer.
\newblock Gendered differences in face recognition accuracy explained by
  hairstyles, makeup, and facial morphology.
\newblock {\em {IEEE} Trans. Inf. Forensics Secur.}, 17:127--137, 2022.

\bibitem{DBLP:conf/btas/BernhardBBF15}
John~S. Bernhard, Jeremiah~R. Barr, Kevin~W. Bowyer, and Patrick~J. Flynn.
\newblock Near-ir to visible light face matching: Effectiveness of
  pre-processing options for commercial matchers.
\newblock In {\em {IEEE} 7th International Conference on Biometrics Theory,
  Applications and Systems}, pages 1--8. {IEEE}, 2015.

\bibitem{DBLP:journals/pami/Best-RowdenJ18}
Lacey Best{-}Rowden and Anil~K. Jain.
\newblock Longitudinal study of automatic face recognition.
\newblock {\em {IEEE} Trans. Pattern Anal. Mach. Intell.}, 40(1):148--162,
  2018.

\bibitem{DBLP:journals/cviu/BeveridgeGPD09}
J.~Ross Beveridge, Geof~H. Givens, P.~Jonathon Phillips, and Bruce~A. Draper.
\newblock Factors that influence algorithm performance in the face recognition
  grand challenge.
\newblock {\em Comput. Vis. Image Underst.}, 113(6):750--762, 2009.

\bibitem{DBLP:conf/fat/BuolamwiniG18}
Joy Buolamwini and Timnit Gebru.
\newblock Gender shades: Intersectional accuracy disparities in commercial
  gender classification.
\newblock In {\em Conference on Fairness, Accountability and Transparency,
  {FAT}}, volume~81 of {\em Proceedings of Machine Learning Research}, pages
  77--91, 2018.

\bibitem{DBLP:conf/cvpr/DengGXZ19}
Jiankang Deng, Jia Guo, Niannan Xue, and Stefanos Zafeiriou.
\newblock Arcface: Additive angular margin loss for deep face recognition.
\newblock In {\em {IEEE} Conference on Computer Vision and Pattern Recognition,
  {CVPR}}, pages 4690--4699. Computer Vision Foundation / {IEEE}, 2019.

\bibitem{DBLP:journals/csr/FarokhiFS16}
Sajad Farokhi, Jan Flusser, and Usman~Ullah Sheikh.
\newblock Near infrared face recognition: {A} literature survey.
\newblock {\em Comput. Sci. Rev.}, 21:1--17, 2016.

\bibitem{GIVENS2013236}
G.H. Givens, J.R. Beveridge, P.J. Phillips, B.~Draper, Y.M. Lui, and D.~Bolme.
\newblock Introduction to face recognition and evaluation of algorithm
  performance.
\newblock {\em Computational Statistics and Data Analysis}, 67:236--247, 2013.

\bibitem{9320939}
Jinyu Gong, Anran Wang, and Weidong Chen.
\newblock Lightcnn: A compact cnn for moving maritime targets detection.
\newblock In {\em 15th IEEE International Conference on Signal Processing
  (ICSP)}, volume~1, pages 321--326, 2020.

\bibitem{42061}
Patrick Grother, George Quinn, and P~Phillips.
\newblock Report on the evaluation of 2d still-image face recognition
  algorithms, 2010-06-17 2010.

\bibitem{DBLP:conf/cvpr/HeZRS16}
Kaiming He, Xiangyu Zhang, Shaoqing Ren, and Jian Sun.
\newblock Deep residual learning for image recognition.
\newblock In {\em {IEEE} Conference on Computer Vision and Pattern Recognition,
  {CVPR}}, pages 770--778. {IEEE} Computer Society, 2016.

\bibitem{DBLP:conf/cvpr/HuangLMW17}
Gao Huang, Zhuang Liu, Laurens van~der Maaten, and Kilian~Q. Weinberger.
\newblock Densely connected convolutional networks.
\newblock In {\em {IEEE} Conference on Computer Vision and Pattern Recognition,
  {CVPR}}, pages 2261--2269. {IEEE} Computer Society, 2017.

\bibitem{8756625}
Isabelle Hupont and Carles Fernández.
\newblock Demogpairs: Quantifying the impact of demographic imbalance in deep
  face recognition.
\newblock In {\em 14th IEEE International Conference on Automatic Face Gesture
  Recognition}, pages 1--7, 2019.

\bibitem{DBLP:conf/mva/KimuraN11}
Yoshikatsu Kimura and Takashi Naito.
\newblock Human skin detection by visible and near-infrared imaging.
\newblock In {\em Proceedings of the {IAPR} Conference on Machine Vision
  Applications {(IAPR}}, pages 503--507, 2011.

\bibitem{DBLP:journals/tifs/KlareBKBJ12}
Brendan Klare, Mark~James Burge, Joshua~C. Klontz, Richard W.~Vorder Bruegge,
  and Anil~K. Jain.
\newblock Face recognition performance: Role of demographic information.
\newblock {\em {IEEE} Trans. Inf. Forensics Secur.}, 7(6):1789--1801, 2012.

\bibitem{DBLP:conf/icmla/KrishnanAR20}
Anoop Krishnan, Ali Almadan, and Ajita Rattani.
\newblock Understanding fairness of gender classification algorithms across
  gender-race groups.
\newblock In {\em 19th {IEEE} International Conference on Machine Learning and
  Applications, {ICMLA}}, pages 1028--1035. {IEEE}, 2020.

\bibitem{9001031}
K.~S. Krishnapriya, Vítor Albiero, Kushal Vangara, Michael~C. King, and
  Kevin~W. Bowyer.
\newblock Issues related to face recognition accuracy varying based on race and
  skin tone.
\newblock {\em IEEE Transactions on Technology and Society}, 1(1):8--20, 2020.

\bibitem{news}
Angellca Marl.
\newblock São paulo subway ordered to suspend use of facial recognition, 2022.

\bibitem{DBLP:journals/tifs/MuhammadWWZS21}
Jawad Muhammad, Yunlong Wang, Caiyong Wang, Kunbo Zhang, and Zhenan Sun.
\newblock Casia-face-africa: {A} large-scale african face image database.
\newblock {\em {IEEE} Trans. Inf. Forensics Secur.}, 16:3634--3646, 2021.

\bibitem{DBLP:journals/corr/abs-2207-10246}
Aakash~Varma Nadimpalli and Ajita Rattani.
\newblock {GBDF:} gender balanced deepfake dataset towards fair deepfake
  detection.
\newblock {\em CoRR}, abs/2207.10246, 2022.

\bibitem{9660182}
Aakash~Varma Nadimpalli, Narsi Reddy, Sreeraj Ramachandran, and Ajita Rattani.
\newblock Harnessing unlabeled data to improve generalization of biometric
  gender and age classifiers.
\newblock In {\em 2021 IEEE Symposium Series on Computational Intelligence
  (SSCI)}, pages 1--7, 2021.

\bibitem{DBLP:conf/fgr/OTooleAPD11}
Alice~J. O'Toole, Xiaobo An, P.~Jonathon Phillips, and Joseph~P. Dunlop.
\newblock Demographic effects on estimates of automatic face recognition
  performance.
\newblock In {\em 9th {IEEE} International Conference on Automatic Face and
  Gesture Recognition}, pages 83--90. {IEEE} Computer Society, 2011.

\bibitem{DBLP:conf/aies/RajiB19}
Inioluwa~Deborah Raji and Joy Buolamwini.
\newblock Actionable auditing: Investigating the impact of publicly naming
  biased performance results of commercial {AI} products.
\newblock In {\em Proceedings of the 2019 {AAAI/ACM} Conference on AI, Ethics,
  and Society}, pages 429--435. {ACM}, 2019.

\bibitem{DBLP:conf/eccv/SaxenaV16}
Shreyas Saxena and Jakob Verbeek.
\newblock Heterogeneous face recognition with cnns.
\newblock In {\em {ECCV} Workshops}, volume 9915 of {\em Lecture Notes in
  Computer Science}, pages 483--491, 2016.

\bibitem{DBLP:journals/ijcv/SelvarajuCDVPB20}
Ramprasaath~R. Selvaraju, Michael Cogswell, Abhishek Das, Ramakrishna Vedantam,
  Devi Parikh, and Dhruv Batra.
\newblock Grad-cam: Visual explanations from deep networks via gradient-based
  localization.
\newblock {\em Int. J. Comput. Vis.}, 128(2):336--359, 2020.

\bibitem{DBLP:conf/aaai/SernaMFCOR20}
Ignacio Serna, Aythami Morales, Julian Fi{\'{e}}rrez, Manuel Cebri{\'{a}}n,
  Nick Obradovich, and Iyad Rahwan.
\newblock Algorithmic discrimination: Formulation and exploration in deep
  learning-based face biometrics.
\newblock In {\em SafeAI@AAAI, Workshop on Artificial Intelligence Safety},
  volume 2560 of {\em {CEUR} Workshop Proceedings}, pages 146--152.
  CEUR-WS.org, 2020.

\bibitem{Siddiqui_2022_CVPR}
Hera Siddiqui, Ajita Rattani, Karl Ricanek, and Twyla Hill.
\newblock An examination of bias of facial analysis based bmi prediction
  models.
\newblock In {\em Proceedings of the IEEE/CVF Conference on Computer Vision and
  Pattern Recognition (CVPR) Workshops}, pages 2926--2935, June 2022.

\bibitem{DBLP:conf/cvpr/Vera-RodriguezB19}
Rub{\'{e}}n Vera{-}Rodr{\'{\i}}guez, Marta Bl{\'{a}}zquez, Aythami Morales,
  Ester Gonzalez{-}Sosa, Jo{\~{a}}o~C. Neves, and Hugo Proen{\c{c}}a.
\newblock Facegenderid: Exploiting gender information in dcnns face recognition
  systems.
\newblock In {\em {IEEE} Conference on Computer Vision and Pattern Recognition
  Workshops}, pages 2254--2260. Computer Vision Foundation / {IEEE}, 2019.

\end{thebibliography}

\end{document}